\documentclass[final]{siamltex}
\usepackage[cmex10]{amsmath}
\usepackage{amssymb}
\usepackage{graphicx}
\usepackage[dvipdfm]{hyperref}
\graphicspath{{figures/}}

 \newcommand{\RR}{\mathbb{R}}

 \newcommand{\T}{{}^\top}

 \newcommand{\bm}[1]{\boldsymbol{#1}}

 \newcommand{\va}{\boldsymbol{a}}

 \newcommand{\vz}{\boldsymbol{z}}
 \newcommand{\vx}{\boldsymbol{x}}
 \newcommand{\vy}{\boldsymbol{y}}
 \newcommand{\vw}{\boldsymbol{w}}

 \newcommand{\valpha}{\boldsymbol{\alpha}}

 \newcommand{\mA}{\boldsymbol{A}}
 
 \newcommand{\mG}{\boldsymbol{G}}
 \newcommand{\mS}{\boldsymbol{S}}
 \newcommand{\mW}{\boldsymbol{W}}
 \newcommand{\mX}{\boldsymbol{X}}
 \newcommand{\mY}{\boldsymbol{Y}}
 \newcommand{\mZ}{\boldsymbol{Z}}
 \newcommand{\mP}{\boldsymbol{P}}
 \newcommand{\mU}{\boldsymbol{U}}
 \newcommand{\mV}{\boldsymbol{V}}
 \newcommand{\mI}{\boldsymbol{I}}

 \newcommand{\mOmega}{\boldsymbol{\Omega}}

 \newcommand{\tG}{\boldsymbol{\mathcal{G}}}
 \newcommand{\tX}{\boldsymbol{\mathcal{X}}}

 \newcommand{\tLambda}{\boldsymbol{\Lambda}}

 \newcommand{\minimize}{\mathop{\rm minimize}}
 \newcommand{\maximize}{\mathop{\rm maximize}}
 \newcommand{\argmin}{\mathop{\rm argmin}}
 
 \newcommand{\subjectto}{\mbox{\rm subject to}}

 \newcommand{\Eqref}[1]{Equation~{\eqref{#1}}}
 
 \newcommand{\Secref}[1]{Section~{\ref{#1}}}
 \newcommand{\Figref}[1]{Fig.~{\ref{#1}}}

\newcommand{\ssum}{\textstyle\sum\limits}

\newcommand{\prox}[1]{{\rm prox}_{#1}}

\newcommand{\nop}[1]{\bar{n}_{\backslash #1}}

\title{Estimation of low-rank tensors via convex
optimization\thanks{This work was supported by MEXT KAKENHI 22700138,
80545583, JST PRESTO, and NTT Communication Science Laboratories.}}
\author{Ryota Tomioka\thanks{Department of Mathematical Informatics, The
University of Tokyo, 7-3-1 Hongo, Bunkyo-ku, Tokyo, 113-8656, Japan
({\tt \{tomioka,kashima\}@mist.i.u-tokyo.ac.jp})
}
\and Kohei Hayashi\thanks{Graduate School of Information Science,
Nara Institute of Science and Technology,
8916-5 Takayama, Ikoma, Nara, 630-0192, Japan
 ({\tt kohei-h@is.naist.jp})
},
\and  Hisashi Kashima$^{\dagger}$}

\begin{document}
\maketitle

\begin{abstract}
In this paper, we propose three approaches for the estimation of the Tucker
 decomposition of multi-way arrays (tensors) from partial observations. All approaches are
 formulated as convex minimization problems. Therefore, the minimum is
 guaranteed to be unique. The proposed approaches can automatically
 estimate the number of factors (rank) through the
 optimization. Thus, there is no need to specify the rank 
 beforehand. The key technique we employ is the trace norm regularization,
 which is a popular approach for the estimation of low-rank matrices.
In addition, we propose a simple heuristic to improve the
 interpretability of the obtained factorization.
 The advantages and disadvantages of three proposed approaches are
 demonstrated through numerical experiments on both synthetic and real
 world datasets. We show that the proposed convex optimization based
 approaches are more accurate in predictive performance, faster, and
 more reliable in recovering a known multilinear structure than
 conventional approaches.
\end{abstract}
\section{Introduction}
Multi-way data analysis have recently become increasingly popular
supported by modern computational power~\cite{KolBad09,SmiBroGel04}. Originally developed in the field of
psychometrics and chemometrics, its applications can now also be found in
signal processing (for example, for independent component
analysis)~\cite{DeLVan04}, neuroscience~\cite{MorHanHerParArn06}, and
data mining~\cite{Mor11}. Decomposition of multi-way
arrays (or tensors) into small number of factors have been one of the
main concerns in multi-way data analysis, because interpreting the original
multi-way data is often impossible. There  are two popular models for
tensor decomposition, namely the Tucker
decomposition~\cite{Tuc66,DeLDeMVan00} and the CANDECOMP/PARAFAC (CP)
decomposition~\cite{CarCha70,Har70}. In both cases, conventionally the
estimation procedures have been formulated as non-convex optimization
problems, which are in general only guaranteed to converge locally and
could potentially suffer from poor local minima. Moreover, a popular
approach for Tucker decomposition known as the higher order orthogonal
iteration (HOOI) may converge to a stationary point that is not even a
local minimizer~\cite{EldSav09}.

Recently, convex formulations for the estimation of low-rank {\em matrix},
which is a special case of tensor, have been intensively studied. After the
pioneering work of Fazel et al.~\cite{FazHinBoy01}, convex optimization
has been used for collaborative filtering~\cite{SreRenJaa05}, multi-task
learning~\cite{ArgEvgPon07}, and classification over
matrices~\cite{TomAih07}.  In addition, there are theoretical developments that (under
some conditions) guarantee
 {\em perfect reconstruction} of a low-rank matrix from partial measurements
via convex estimation~\cite{CanRec09,RecFazPar10}. 
The key idea here is to replace the rank of a
matrix (a non-convex function) by the so-called trace norm (also known
as the nuclear norm) of the matrix.
One goal of this paper is to extend
the trace-norm regularization for more than two dimensions.
There have recently been related work by Liu et al.~\cite{LiuMusWonYe09}
and Signoretto et al.~\cite{SigDeLSuy10}, which correspond to one of the
proposed approaches in the current paper.

In this paper, we propose three formulations for the estimation of low
rank tensors. The first approach is called ``as a matrix'' and estimates
the low-rank matrix that is obtained by {\em unfolding} (or
matricizing) the tensor to be estimated; thus this approach basically
treats the unknown tensor as a matrix and only works if the tensor is
low-rank in the mode used for the estimation. The second approach called
``constraint'' extends the first approach by incorporating the trace
norm penalties with respect to all modes simultaneously. Therefore,
there is no arbitrariness in choosing a single mode to work
with. However, all modes being simultaneously low-rank might be a strong
assumption. The third approach called ``mixture'' relaxes the assumption
by using a mixture of $K$ tensors, where $K$ is the number of modes of
the tensor.  Each tensor is regularized to be low-rank in each mode.

We apply the above three approaches to the reconstruction of partially
observed tensors. In both synthetic and real-world datasets, we show the
superior predictive performance of the proposed approaches against
conventional expectation maximization (EM) based estimation of Tucker
decomposition model. We also demonstrate the effectiveness of a
heuristic to improve the interpretability of the core tensor obtained
by the proposed approaches on the amino acid fluorescence dataset.

This paper is structured as follows. In the next section, we first
review the matrix rank 
and its relation to the trace norm. Then we review the definition of
tensor mode-$k$ rank, which suggests that a low rank tensor {\em is} a low
rank matrix when appropriately unfolded. In \Secref{sec:method}, we
propose three approaches to extend the trace-norm regularization for the
estimation of 
low-rank tensors. In \Secref{sec:opt}, we show that the optimization
problems associated to the proposed extensions can be solved efficiently
by the
alternating direction method of multipliers~\cite{GabMer76}. In \Secref{sec:exp}, we
show through numerical experiments that one of
the proposed approaches can recover a partly observed low-rank tensor
almost perfectly from smaller fraction of observations compared to the
conventional EM-based Tucker decomposition algorithm. The proposed
algorithm shows a sharp threshold behaviour from a poor fit to a nearly
perfect fit; we numerically show that the fraction of samples at the
threshold is roughly proportional to the sum of the $k$-ranks of the
underlying tensor when the tensor dimension is fixed. Finally we
summarize the paper in \Secref{sec:summary}. Earlier version of this
manuscript appeared in NIPS2010 workshop ``Tensors, Kernels, and Machine Learning''.

\section{Low rank matrix and tensor}
In this section, we first discuss the connection between the rank of a
matrix and the trace-norm regularization. Then we review the CP and the
Tucker decomposition and the notions of tensor rank connected to them.
\subsection{Rank of a matrix and the trace norm}
The rank $r$ of an $R\times C$ matrix $\mX$ can be defined as the number
of nonzero singular values of $\mX$. Here, the singular-value
decomposition (SVD) of $\mX$ is written as follows:
\begin{align}
\label{eq:svd}
 \mX&=\mU{\rm diag}(\sigma_1(\mX),\sigma_2(\mX),\ldots,\sigma_r(\mX))\mV\T,
\end{align}
where $\mU\in\RR^{R\times r}$ and $\mV\in\RR^{C\times r}$ are orthogonal
matrices, and $\sigma_j(\mX)$ is the $j$th largest singular-value of $\mX$.
The matrix $\mX$ is called {\em low-rank} if the rank $r$ is less than
$\min(R,C)$. Unfortunately, the rank of a matrix is a nonconvex
function, and the direct minimization of rank or solving a
rank-constrained problem is an NP-hard problem~\cite{RecFazPar10}.

\begin{figure}[tb]
 \begin{center}
  \includegraphics[width=.7\textwidth]{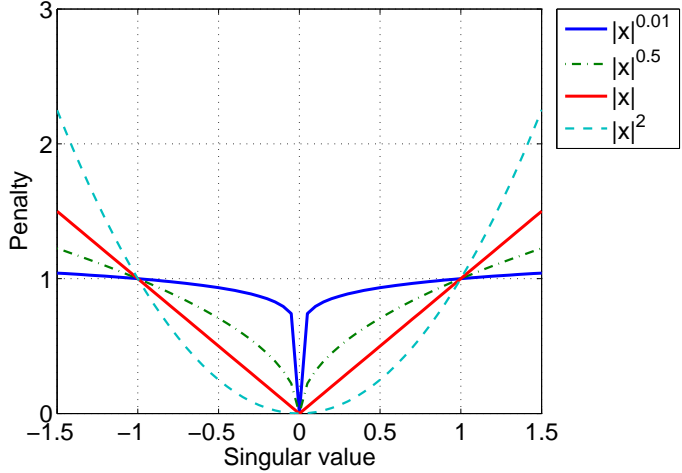}
  \caption{Penalty functions $|x|^p$ over one singular value $x$ are schematically
  illustrated for various $p$. The absolute penalty function $|x|$ is
  the tightest convex lower bound of the rank ($p\rightarrow 0$) in the
  interval $[-1,1]$.}
  \label{fig:norms}
 \end{center}
\end{figure}

The trace norm is known to be the tightest convex lower bound of
matrix rank~\cite{RecFazPar10} (see \Figref{fig:norms}) and is defined
as the linear sum of singular values as follows:
\begin{align*}
 \|\mX\|_{\ast}&=\sum_{j=1}^{r}\sigma_j(\mX).
\end{align*}
Intuitively, the trace norm plays the role of the $\ell_1$-norm in the subset
selection problem~\cite{Tib96}, for the estimation of low-rank
matrix\footnote{Note however that the absolute value is not taken here
because singular value is defined to be positive.}. 
The convexity of the above function follows from the
fact that it is the dual norm of the spectral norm
$\|\cdot\|$ (see
\cite[Section A.1.6]{BoydBook}). Since it is a norm,
the trace norm $\|\cdot\|_{\ast}$ is a convex function.
The non-differentiability of the trace norm at the origin
promotes many singular values of $\mX$ to be zero when used as a
regularization term. In fact, the following minimization problem
has an analytic solution known as the spectral soft-thresholding operator (see \cite{CaiCanShe08}):
\begin{align}
\label{eq:minprox}
 \argmin_{\mX}\quad&\frac{1}{2}\|\mX-\mY\|^2_{\rm
 Fro}+\lambda\|\mX\|_{\ast},
\end{align}
where $\|\cdot\|_{\rm Fro}$ is the Frobenius norm, and $\lambda>0$ is a
regularization constant. The spectral soft-thresholding operation
can be considered as a shrinkage operation on the singular values and is
defined as follows:
\begin{align}
\label{eq:softth}
 &\prox{\lambda}^{\rm tr}(\mY)=\mU\max(\mS-\lambda,0)\mV\T,
\end{align}
where $\mY=\mU\mS\mV\T$ is the SVD of the input matrix $\mY$, and the
$\max$ operation is taken element-wise. We can see that the spectral soft-thresholding operation truncates the singular-values of the input matrix
$\mY$ smaller than $\lambda$ to zero, thus the resulting matrix $\mX$ is
usually low-rank. See also \cite{TomSuzSug11} for the derivation. 

For the recovery of partially observed low-rank matrix, 
some theoretical guarantees have recently been developed. Cand\`es and
Recht~\cite{CanRec09} showed that in the noiseless case,
$O(n^{6/5}r\log(n))$ samples are enough to perfectly recover the matrix
under uniform sampling if the rank $r$ is not too large, where $n=\max(R,C)$.

\subsection{Rank of a tensor}
For higher order tensors, there are several definitions of rank. 
Let $\tX\in\RR^{n_1\times n_2\times \cdots\times n_K}$ be a $K$-way tensor.
The rank of a tensor (see ~\cite{Kru89}) is defined as the minimum number $r$ of
components required for rank-one decomposition of a given tensor $\tX$ in analogy to
SVD as follows:
\begin{align}
 \tX&=\sum_{j=1}^{r}\lambda_j\va_j^{(1)}\circ\va_j^{(2)}\circ\cdots\circ\va_j^{(K)},\nonumber\\
\label{eq:cp}
&=\tLambda\times_{1}\mA^{(1)}\times_{2}\mA^{(2)}\cdots\times_{K}\mA^{(K)}
\end{align}
where $\circ$ denotes the outer product,
$\tLambda\in\RR^{r\times\cdots\times r}$ denotes a $K$-way diagonal
matrix whose $(j,j,j)$th element is $\lambda_j$, and $\times_{k}$ denotes the
$k$-mode matrix product (see Kolda \& Bader~\cite{KolBad09}); in
addition, we define $\mA^{(k)}=[\va_1^{(k)},\ldots,\va_r^{(k)}]$. The above decomposition model is called
CANDECOMP~\cite{CarCha70} or PARAFAC~\cite{Har70}.  It is worth noticing
that finding the above decomposition with the minimum $r$ is a hard problem; thus there is no
straightforward algorithm for computing the rank for higher-order
tensors~\cite{Kru89}.

We consider instead the mode-$k$ rank of tensors, which is the foundation of
the Tucker decomposition~\cite{Tuc66,DeLDeMVan00}. 
The mode-$k$ rank of $\tX$, denoted ${\rm rank}_k(\tX)$, 
is the dimensionality of the space spanned by the mode-$k$ fibers of
$\tX$. In other words, the mode-$k$ rank of $\tX$ is the rank of the
mode-$k$ unfolding $\mX_{(k)}$ of $\tX$.  
The
mode-$k$ unfolding $\mX_{(k)}$ is the $n_k\times\nop{k}$ ($\nop{k}:=\prod_{k'\neq k}n_{k'}$)
matrix obtained by concatenating the mode-$k$ fibers of $\tX$ as column
vectors. In MATLAB this can be obtained as follows:
\begin{center}
\verb| X=permute(X,[k:K,1:k-1]); X=X(:,:);|
\end{center}
where the order of dimensions other than the first dimension $k$ is not important
as long as we use a consistent definition. We say that a $K$-way tensor $\tX$ is
rank-$(r_1,\ldots,r_K)$ if the mode-$k$ rank of $\tX$ is $r_k$
($k=1,\ldots,K$). 
Unlike the rank of the tensor, mode-k rank is clearly computable; the computation of the mode-$k$
ranks of a tensor boils down to the computation the rank of $K$ matrices.

%

A rank-$(r_1,\ldots,r_K)$ tensor $\tX\in\RR^{n_1\times\cdots\times n_K}$ can be written as
\begin{align}
\label{eq:tucker}
 \tX&=\tG\times_{1}\mU_1\times_{2}\mU_2\cdots\times_{K}\mU_K,
\end{align}
where $\tG\in\RR^{r_1\times\cdots\times r_K}$ is called a {\em core
tensor}, and $\mU_k\in\RR^{n_k\times r_k}$ ($n=1,\ldots,K$) are left
singular-vectors from the SVD of the mode-$k$ unfolding of $\tX$. The
above decomposition is called the Tucker decomposition~\cite{Tuc66,KolBad09}.

The definition of a low-rank tensor (in the sense of Tucker
decomposition) implies that a low-rank tensor {\em is} a low-rank matrix
when unfolded appropriately. In order to see this, we recall that for the Tucker
model~\eqref{eq:tucker}, the mode-$k$ unfolding of $\tX$ can be written
as follows (see e.g., \cite{KolBad09}):
\begin{align*}
 \mX_{(k)}&=\mU_k\mG_{(k)}\left(\mU_{k-1}\otimes\cdots\otimes\mU_{1}\otimes\mU_{K}\otimes\cdots\otimes\mU_{k+1}\right)\T.
\end{align*}
Therefore, if the tensor $\tX$ is low-rank in the $k$th mode (i.e.,
$r_k<\min(n_k,\nop{k})$), its unfolding is a low-rank
matrix. Conversely, if $\mX_{(k)}$ is a low-rank matrix (i.e.,
$\mX_{(k)}=\mU\mS\mV\T$), we can set $\mU_k=\mU$, $\mG_{(k)}=\mS\mV\T$,
and other Tucker factors $\mU_{k'}$ $(k'\neq k)$ as identity matrices,
and we obtain a Tucker decomposition \eqref{eq:tucker}.

Note that if a given tensor $\tX$ can be written in the form of CP
decomposition~\eqref{eq:cp} with rank $r$, 
we can always find a rank-$(r,r,\ldots,r)$
Tucker decomposition~\eqref{eq:tucker} by ortho-normalizing each factor
$\mA^{(k)}$ in \Eqref{eq:cp}. Therefore, the Tucker
decomposition is more general than the CP decomposition.

However, since the core tensor $\tG$ that corresponds to singular-values in the
matrix case (see \Eqref{eq:svd}) is not diagonal in general, it is not
straightforward to generalize the trace norm from matrices to tensors.

\section{Three strategies to extend the trace-norm regularization to tensors}
\label{sec:method}
In this section, we first consider a given tensor as a matrix by
unfolding it at a given mode $k$ and propose to
minimize the trace norm of the unfolding $\mX_{(k)}$. Next, we extend this
to the minimization of the weighted sum of the trace norms of the
unfoldings. Finally, relaxing the condition that the tensor is
{\em  jointly} low-rank in every mode in the second approach, we propose
a mixture approach. For solving the optimization problems, we use the
alternating direction method of multipliers (ADMM)~\cite{GabMer76} (also known
as the split Bregman iteration~\cite{GolOsh09}). The
optimization algorithms are discussed in \Secref{sec:opt}.

\subsection{Tensor as a matrix}
If we assume that the tensor we wish to estimate is (at least) low-rank
in the $k$th mode, we can convert the tensor estimation problem into a
matrix estimation problem. Extending the minimization problem
\eqref{eq:minprox} to accommodate missing entries we have the following
optimization problem for the reconstruction of partially observed tensor:
\begin{align}
\label{eq:asmatrix}
 \minimize_{\tX\in\RR^{n_1\times\cdots\times n_K}}\qquad
 &\frac{1}{2\lambda}\|\Omega(\tX)-\vy\|^2+\|\mX_{(k)}\|_{\ast}, 
\end{align}
where $\mX_{(k)}$ is the mode-$k$ unfolding of $\tX$, $\vy\in\RR^{M}$ is
the vector of observations, and
$\Omega:\RR^{n_1\times\cdots\times n_K}\rightarrow \RR^M$ is a linear
operator that reshapes the prespecified  elements
of the input tensor into an $M$ dimensional vector; $M$ is the number of
observations. In \Eqref{eq:asmatrix}, the regularization constant
$\lambda>0$ is moved to the denominator of the loss term from the
numerator of the regularization term in \Eqref{eq:minprox}; this
equivalent reformulation allows us to consider the noiseless case
$(\lambda\rightarrow 0)$ in the same framework. Note that $\lambda$ can
also be interpreted as the variance of the Gaussian observation noise model.

Since the estimation procedure \eqref{eq:asmatrix} is essentially
an estimation of a low-rank matrix $\mX_{(k)}$, we know that
in the noiseless case $O(\tilde{n}_k^{6/5}r_k\log (\tilde{n}_k))$
samples are enough to perfectly recover the unknown true tensor
$\tX^\ast$, where $r_k={\rm rank}_k(\tX^{\ast})$ and
$\tilde{n}_k=\max(n_k, \nop{k})$, if the rank $r_k$ is
not too high~\cite{CanRec09}. This holds regardless of whether the
unknown tensor $\tX$ is low-rank in other modes $k'\neq k$. 
Therefore, when we can estimate the mode-$k$ unfolding of $\tX^{\ast}$ 
perfectly, we can also recover the whole $\tX^{\ast}$ perfectly,
including the ranks of the modes we did not use during the estimation.

However, the success of the above procedure is conditioned on
the choice of the mode to unfold the tensor. 
If we choose a mode with a large rank, even if there are other modes
with smaller ranks, we cannot hope to recover the tensor from a small
number of samples.

Various advanced
methods~\cite{TomSuzSug11,TomSuzSugKas10,LinCheWuMa09,JiYe09} 
for the estimation of low-rank matrices can be used for solving the
minimization problem~\eqref{eq:asmatrix}. Here we use ADMM 
to keep the presentation concise; see \Secref{sec:opt} for the details.

\subsection{Constrained optimization of low rank tensors}
In order to exploit the rank deficiency of more than one mode, 
it is natural to consider the following extension of
 the estimation procedure~\eqref{eq:asmatrix}
\begin{align}
\label{eq:const-basic}
\minimize_{\tX\in\RR^{n_1\times\cdots\times n_K}}
\qquad &\frac{1}{2\lambda}\|\Omega(\tX)-\vy\|^2+\sum_{k=1}^K\gamma_k\|\mX_{(k)}\|_{\ast}.
\end{align}
This is a convex optimization problem, because it can be reformulated as
follows:
\begin{align}
\label{eq:const-obj}
 \minimize_{\vx,\mZ_1,\ldots,\mZ_K}\qquad&\frac{1}{2\lambda}\|\mOmega\vx-\vy\|^2+
\sum_{k=1}^K\gamma_k\|\mZ_k\|_{\ast},\\
\label{eq:const-const}
\subjectto\qquad &\mP_k\vx=\vz_k\quad(k=1,\ldots,K),
\end{align}
where $\vx\in\RR^{N}$ is the vectorization of $\tX$
($N=\prod_{k=1}^Kn_k$), $\mP_k$ is the matrix representation of mode-$k$
unfolding (note that $\mP_k$ is a permutation matrix; thus
$\mP_k\T\mP_k=\mI_N$), $\mZ_k\in\RR^{n_k\times \nop{k}}$ is an auxiliary
matrix of the same size as the mode-$k$ unfolding of $\tX$, and $\vz_k$ is the
vectorization of $\mZ_k$. With a slight abuse of notation
$\mOmega\in\RR^{M\times N}$ denotes the observation operator as a matrix.

This approach was considered earlier by Liu et al.~\cite{LiuMusWonYe09} and
Signoretto et al.~\cite{SigDeLSuy10}. Liu et al. relaxed the
constraints~\eqref{eq:const-const} into penalty terms, therefore the
factors obtained as the left singular vectors of $\mZ_k$ does not equal
the factors of the Tucker decomposition of $\tX$. Signoretto et
al. have discussed the general Shatten-$\{p,q\}$ norms for tensors and
the relationship between the regularization term in
\Eqref{eq:const-basic} with $\gamma_k=1/K$  (which corresponds to Shatten-$\{1,1\}$ norm)  and the function 
$\frac{1}{K}\sum_{k=1}^K{\rm  rank}_k(\tX)$.



\subsection{Mixture of low-rank tensors}
The optimization problem \eqref{eq:const-obj} penalizes every mode of
the tensor $\tX$ to be {\em jointly} low-rank, which might be too strict
to be satisfied in practice. Thus we propose to predict instead with a
mixture of $K$ tensors; each mixture component is regularized by the
trace norm to be low-rank in each mode. More specifically, we solve the following minimization problem:
\begin{align}
 \label{eq:mixture}
\minimize_{\mZ_1,\ldots,\mZ_K}\qquad&\frac{1}{2\lambda}\left\|\mOmega\left(\sum\nolimits_{k=1}^K\mP_k\T\vz_k\right)-\vy\right\|^2+\sum_{k=1}^K\gamma_k\|\mZ_k\|_{\ast}.
\end{align}
Note that when $\vz_k=\frac{1}{K}\mP_k\vx$ for all $k=1,\ldots,K$, the problem
\eqref{eq:mixture} reduces to the problem \eqref{eq:const-obj} with $\gamma_k'=\gamma_k/K$.

\subsection{Interpretation}
\label{sec:interpretation}
All three proposed approaches inherit the {\em lack of uniqueness} of
the factors from the conventional Tucker
decomposition~\cite{KolBad09}. Some heuristics to improve the
interpretability of the core tensor $\tG$ are proposed and implemented in
the $N$-way toolbox~\cite{AndBro00}. However, these approaches are all
restricted to orthogonal transformations. Here we present another
simple heuristic, which is to apply PARAFAC decomposition on the core
tensor $\tG$. This approach has the following advantages over applying
PARAFAC directly to the original data. First, the dimensionality of the
core tensor $(r_1,\dots,r_K)$ is 
automatically obtained from the proposed algorithms. Therefore, the range
of the number of PARAFAC components that we need to look for is much
narrower than applying PARAFAC directly to the original data. Second,
the PARAFAC problem does not need to take care of missing entries. In
other words, we can separate the prediction problem and the
interpretation problem, which are separately tackled by the proposed
algorithms and PARAFAC, respectively. Finally, empirically the proposed
heuristic seems to be more robust in recovering the underlying factors compared to
applying PARAFAC directory when the rank is misspecified (see \Secref{sec:exp-amino}).

More precisely, let us consider the second ``Constraint'' approach. Let
$\mU_1\ldots,\mU_K$ be the left singular vectors of the auxiliary
variables $\mZ_1,\ldots,\mZ_K$. From \Eqref{eq:tucker} we can obtain the
core tensor $\tG$ as follows:
\begin{align*}
 \tG&=\tX\times_{1}\mU_1\T\times_{2}\mU_2\T\cdots\times_{K}\mU_{K}\T.
\end{align*}
Let $\mA^{(1)},\ldots,\mA^{(K)}$ be the factors obtained by the PARAFAC
decomposition of $\tG$ as follows:
\begin{align*}
 \tG&=\tLambda\times_{1}\mA^{(1)}\times_{2}\mA^{(2)}\cdots\times_{K}\mA^{(K)}.
\end{align*}
Therefore, we have the following decomposition
\begin{align}
\label{eq:parafac-factors}
 \tX&=\tLambda\times_{1}(\mU_1\mA^{(1)})\times_{2}(\mU_{2}\mA^{(2)})\cdots\times_{K}(\mU_{K}\mA^{(K)}),
\end{align}
which gives the $k$th factor as $\mU_{k}\mA^{(k)}$.
\section{Optimization}
\label{sec:opt}
In this section, we describe the optimization algorithms based on the
alternating direction method of multipliers (ADMM) for the
problems~\eqref{eq:asmatrix}, \eqref{eq:const-obj}, and
\eqref{eq:mixture}.

\subsection{ADMM}
The alternating direction method of multipliers~\cite{GabMer76} (see
also \cite{BoyParChuPelEck11}) can be considered as an
approximation of the method of multipliers~\cite{Pow69,Hes69} (see also \cite{Ber82,NocWri99}). The
method of multipliers generates a sequence of primal variables
$(\vx^{t},\vz^{t})$ and multipliers $\valpha^{t}$ by iteratively minimizing the so
called  augmented Lagrangian (AL) function with respect to the primal variables
$(\vx^{t},\vz^{t})$ and updating the multiplier vector $\valpha^{t}$. Let us consider
the following linear equality constrained minimization problem:
\begin{align}
\label{eq:gen-obj}
 \minimize_{\vx\in\RR^{n},\vz\in\RR^{m}} \quad &f(\vx)+g(\vz),\\
\label{eq:gen-const}
\subjectto \quad&\mA\vx=\vz,
\end{align}
where $f$ and $g$ are both convex functions. The AL function $L_{\eta}(\vx,\vz,\valpha)$ of the above minimization problem is written as follows:
\begin{align*}
L_{\eta}(\vx,\vz,\valpha) &=f(\vx)+g(\vz)+\valpha\T(\mA\vx-\vz)+\frac{\eta}{2}\|\mA\vx-\vz\|^2,
\end{align*}
where $\valpha\in\RR^m$ is the Lagrangian multiplier vector. Note that
when $\eta=0$, the AL function reduces to the ordinary Lagrangian
function. Intuitively, the additional penalty term enforces the equality
constraint to be satisfied. However, different from the penalty method
(which was used in \cite{LiuMusWonYe09}), there is no need to increase
the penalty parameter $\eta$ very large, which usually makes the
problem poorly conditioned.

The original method of multipliers performs minimization of the AL
function with respect to $\vx$ and $\vz$ {\em jointly} followed by a
multiplier update as follows:
\begin{align}
\label{eq:alminimize}
 (\vx^{t+1},\vz^{t+1})&=\argmin_{\vx\in\RR^n,\vz\in\RR^{m}}
 L_{\eta}(\vx,\vz,\valpha^t),\\
\label{eq:alupdate}
\valpha^{t+1}&=\valpha^{t}+\eta(\mA\vx^{t+1}-\vz^{t+1}).
\end{align}
Intuitively speaking, the multiplier is updated proportionally to the
violation of the equality constraint~\eqref{eq:gen-const}. In this
sense, $\eta$ can also be regarded as a step-size parameter.
Under fairly
mild conditions, the above method converges
super-linearly to a solution of the minimization
problem~\eqref{eq:gen-obj}; see \cite{Roc76b,TomSuzSug09}. However, the
joint minimization of the AL function~\eqref{eq:alminimize} is often
hard (see \cite{TomSuzSug09} for an exception).

The ADMM decouples the minimization with respect to $\vx$ and $\vz$ as
follows:
\begin{align}
\label{eq:admm-x}
 \vx^{t+1}&=\argmin_{\vx\in\RR^n} L_{\eta}(\vx,\vz^{t},\valpha^{t}),\\
\label{eq:admm-z}
 \vz^{t+1}&=\argmin_{\vz\in\RR^m} L_{\eta}(\vx^{t+1},\vz,\valpha^{t}),\\
\label{eq:admm-a}
 \valpha^{t+1}&=\valpha^{t}+\eta(\mA\vx^{t+1}-\vz^{t+1}).
\end{align}
Note that the new value of $\vx^{t+1}$ obtained in the first line is
used in the update of $\vz^{t+1}$ in the second line. The multiplier
update step is identical to that of the ordinary method of
multipliers~\eqref{eq:alupdate}.  It can be shown that the above algorithm is
an application of firmly nonexpansive mapping and that it converges to a
solution of the original problem~\eqref{eq:gen-obj}. Surprisingly, this
is true for any positive penalty parameter $\eta$~\cite{LioMer79}. This
is in contrast to the fact that a related approach called
forward-backward splitting~\cite{LioMer79} (which was used in
\cite{SigDeLSuy10}) converges only when the step-size parameter $\eta$ is chosen appropriately.

\subsection{Stopping criterion}
As a stopping criterion for terminating the above ADMM algorithm, we
employ the relative duality gap criterion; that is, we stop the algorithm
when the current primal objective value $p(\vx,\vz):=f(\vx)+g(\vz)$ and
the largest dual objective value $\max_{t'=1,\ldots,t}d(\tilde{\valpha}^{t'})$ obtained in the past satisfies the following
equality
\begin{align}
\label{eq:criterion}
 (p(\vx^t,\vz^t)-\max_{t'=1,\ldots,t}d(\tilde{\valpha}^{t'}))/p(\vx^t,\vz^t)&< \epsilon.
\end{align}
Note that the multiplier vector $\valpha^t$ computed in
\Eqref{eq:admm-a} cannot be directly used in the computation of the dual
objective value, because typically $\valpha^{t}$ violates the dual
constraints. See Appendix~\ref{sec:gap} for the details.

The reason we use the duality gap is that the criterion is
invariant to the scale of the observed entries $\vy$ and the size of the
problem $N$.

\subsection{ADMM for the ``As a Matrix'' approach}

We consider the following constrained reformulation of
problem~\eqref{eq:asmatrix}
\begin{align}
 \label{eq:mat-const}
&\minimize_{\vx\in\RR^{N},\mZ\in\RR^{n_k\times\nop{k}}}\quad
 \frac{1}{2\lambda}\|\mOmega\vx-\vy\|^2+\|\mZ\|_{\ast},\quad\subjectto\quad\mP_k\vx=\vz,
\end{align}
where $\vx\in\RR^{N}$ is a vectorization of $\tX$, $\mZ\in\RR^{n_k\times \nop{k}}$ is an
auxiliary variable that corresponds to the mode-$k$ unfolding of $\tX$,
and $\vz\in\RR^{N}$ is the vectorization of $\mZ$.
The AL function of the above constrained
minimization problem can be written as follows:
\begin{align}
\label{eq:alfunc1}
 L_{\eta}(\vx,\mZ,\valpha)&=\frac{1}{2\lambda}\|\mOmega\vx-\vy\|^2+\|\mZ\|_{\ast}+\eta\valpha\T(\mP_k\vx-\vz)+\frac{\eta}{2}\|\mP_k\vx-\vz\|^2,
\end{align}
where $\valpha\in\RR^{N}$ is the Lagrangian multiplier vector that
corresponds to the constraint $\mP_k\vx=\vz$. Note that we rescaled the
Lagrangian multiplier vector $\valpha$  by the factor $\eta$ for the
sake of notational simplicity.

Starting from an initial point $(\vx^{0},\mZ^{0},\valpha^{0})$, we apply
the ADMM explained in the previous section to the AL
function~\eqref{eq:alfunc1}. 
All the steps \eqref{eq:admm-x}--\eqref{eq:admm-a} can be implemented in
closed forms. First, 
minimization with respect to $\vx$ yields,
\begin{align}
\label{eq:mat-x-update}
 \vx^{t+1}&=\left(\mOmega\T\vy+\lambda\eta\mP_k\T(\vz^{t}-\valpha^{t})\right)./(\bm{1}_\Omega+\lambda\eta\bm{1}_N),
\end{align}
where $\bm{1}_{\Omega}$ is an $N$-dimensional vector that has one for
observed elements and zero otherwise; $\bm{1}_N$ is an $N$-dimensional
 vector filled with ones; $./$ denotes element-wise division. Note that
 when $\lambda\rightarrow 0$ (no observational noise), the above
 expression can be simplified as follows:
\begin{align}
\label{eq:mat-x-update-lambda=0}
 x^{t+1}_i&=
\begin{cases}
( \mOmega\T\vy)_i, & i\in\Omega,\\
(\mP_k\T(\vz^t-\valpha^t))_i, & i\notin\Omega
\end{cases} \quad (i=1,\ldots,N).
\end{align}
Here the observed entries of $\vx$ are overwritten by the observed
values $\vy$ and the unobserved entries are filled with the mode-$k$
tensorization of the current prediction~$\vz^{t}-\valpha^{t}$. In the general
case~\eqref{eq:mat-x-update}, the
predicted values also affect the observed entries. The primal variable
$\vx^t$ and the auxiliary variable $\vz^t$ becomes closer and closer as the
optimization proceeds. This means that eventually the multiplier
vector $\valpha^t$ takes non-zero values only on the observed entries when
$\lambda\rightarrow 0$. 

Next, the minimization with respect to $\mZ$ yields,
\begin{align*}
\mZ^{t+1}=\prox{1/\eta}^{\rm tr}\left(\mP_k\vx^{t+1}+\valpha^t\right),
\end{align*}
where $\prox{1/\eta}^{\rm tr}$ is the spectral soft-threshold operation~\eqref{eq:softth} in which the argument
$\mP_k\vx^{t+1}+\valpha^t$ is considered as a
$n_k\times\nop{k}$ matrix. 

The last step is the multiplier update~\eqref{eq:admm-a}, which can be
written as follows:
\begin{align}
\label{eq:mat-alpha-update}
\valpha^{t+1}&=\valpha^{t}+\left(\mP_k\vx^{t+1}-\vz^{t+1}\right).
\end{align}
Note that the step-size parameter $\eta$ does not appear in
\eqref{eq:mat-alpha-update} due to the rescaling of $\valpha$ in \eqref{eq:alfunc1}.

The speed of convergence of the algorithm mildly depends on the choice
of the step-size $\eta$. Here as a guideline to choose $\eta$, we
require that the algorithm is invariant to scalar multiplication of the
objective~\eqref{eq:mat-const}. More precisely, when the input $\vy$ and
the regularization constant $\lambda$  are both multiplied by a constant
$c$, the solution of the minimization~\eqref{eq:mat-const} (or
\eqref{eq:asmatrix}) should remain essentially the same as the original
problem, except that the solution 
$\vx$ is also multiplied by the constant $c$. In order to make the
algorithm (see \eqref{eq:mat-x-update}-\eqref{eq:mat-alpha-update})
follow the same path (except that $\vx^t$, $\vz^t$, and $\valpha^t$ are
all multiplied by $c$), we need to scale $\eta$ inversely proportional
to $c$. We can also see this in the AL function~\eqref{eq:alfunc1}; in fact, the first two terms scale linearly
to $c$, and also the last two terms scale linearly if $\eta$ scales
inversely to $c$.  Therefore we choose $\eta$ as
$\eta=\eta_0/{\rm std}(\vy)$, where $\eta_0$ is a constant and ${\rm
std}(\vy)$ is the standard deviation of the observed values $\vy$.


\subsection{ADMM for the ``Constraint'' approach}
The AL function of the constrained minimization problem
\eqref{eq:const-obj}-\eqref{eq:const-const} can be written as follows:
\begin{align*}
 L_{\eta}(\vx,\{\mZ_k\}_{k=1}^K,\{\valpha_k\}_{k=1}^K)&=\frac{1}{2\lambda}\|\mOmega\vx-\vy\|^2+\sum_{k=1}^K\gamma_k\|\mZ_k\|_{\ast}\\
&\qquad+\sum_{k=1}^K\left(\eta\valpha_k\T(\mP_k\vx-\vz_k)+\frac{\eta}{2}\|\mP_k\vx-\vz_k\|^2\right).
\end{align*}
Note that we rescaled the multiplier vector $\valpha$ by the factor
$\eta$ as in the previous subsection.

Starting from an initial point
$(\vx^0,\{\mZ_k^0\}_{k=1}^K,\{\valpha_k^0\}_{k=1}^K)$, we take similar
steps as in \eqref{eq:mat-x-update}-\eqref{eq:mat-alpha-update}
except that the last two steps are performed for all
$k=1,\ldots,K$. That is,
\begin{align}
 \label{eq:const-x-update}
 \vx^{t+1}&=\left(\mOmega\T\vy+\lambda\eta\sum\nolimits_{k=1}^K\mP_k\T(\vz_k^{t}-\valpha_k^{t})\right)./(\bm{1}_\Omega+\lambda\eta  K\bm{1}_N),\\
 \mZ_k^{t+1}&=\prox{\gamma_k/\eta}^{\rm
 tr}\left(\mP_k\vx^{t+1}+\valpha_k^t\right)\qquad(k=1,\ldots,K),\\
 \valpha_k^{t+1} &=\valpha_k^{t}+(\mP_k\vx^{t+1}-\vz_k^{t+1})\qquad(k=1,\ldots,K).
\end{align}

By considering the scale invariance of the algorithm, we choose the
step-size $\eta$ as $\eta=\eta_0/{\rm std}(\vy)$  as in the previous
subsection.


\subsection{ADMM for the ``Mixture'' approach}
\label{sec:admm-mixture}
We consider the following dual problem of the mixture formulation~\eqref{eq:mixture}:
\begin{align}
\label{eq:mix-dual}
 \minimize_{\valpha\in\RR^M,\mW_k\in\RR^{n_k\times \nop{k}}}\quad
& \frac{\lambda}{2}\|\valpha\|^2-\valpha\T\vy+\sum_{k=1}^K\delta_{\gamma_k}(\mW_k),\\
\subjectto\quad &\vw_k=\mP_k\mOmega\T\valpha\qquad(k=1,\ldots,K),\nonumber
\end{align}
where $\valpha\in\RR^{M}$ is a dual vector; $\mW_k\in\RR^{n_k\times
\nop{k}}$ is an auxiliary variable that corresponds to the mode-$k$
unfolding of $\mOmega\T\valpha$, and $\vw_k\in\RR^{N}$ is the
vectorization of $\mW_k$; the indicator function $\delta_{\lambda}$ is
defined as $\delta_{\lambda}(\mW)=0$, if 
$\|\mW\|\leq\lambda$, and $\delta_{\lambda}(\mW)=+\infty$, otherwise,
where $\|\cdot\|$ is the spectral norm (maximum singular-value of a matrix).

The AL function for the problem \eqref{eq:mix-dual} can be written as
follows:
\begin{align*}
 L_{\eta}(\valpha,\{\mW_k\}_{k=1}^K,\{\vz_k\}_{k=1}^K)&=\frac{\lambda}{2}\|\valpha\|^2-\valpha\T\vy+\sum_{k=1}^{K}\delta_{\gamma_k}(\mW_k)\\
&\qquad+\sum_{k=1}^K\left(\vz_k\T(\mP_k\mOmega\T\valpha-\vw_k)+\frac{\eta}{2}\|\mP_k\mOmega\T\valpha-\vw_k\|^2\right)
\end{align*}

Similar to the previous two algorithms, we start from an initial point
$(\valpha^0,\{\mW_k^0\}_{k=1}^K,\{\vz_k^0\}_{k=1}^K)$, and compute the
following steps:
\begin{align}
 \valpha^{t+1}&=\argmin_{\valpha}L_{\eta}(\valpha,\{\mW_k^t\}_{k=1}^K,\{\vz_k^t\}_{k=1}^K)\nonumber\\
 \mW_k^{t+1}&=\argmin_{\mW_k}L_{\eta}(\valpha^{t+1},\{\mW_k\}_{k=1}^K,\{\vz_k^t\}_{k=1}^K)\nonumber\\
\label{eq:mix-x-update-pre}
\vz_k^{t+1}&=\vz_k^{t}+\eta(\mP_k\mOmega\T\valpha^{t+1}-\vw_k^{t+1}).
\end{align}
The above steps can be computed in closed forms. In fact,
\begin{align}
 \label{eq:mix-alpha-update}
\valpha^{t+1}&=\frac{1}{\lambda+\eta K}\left(\vy-\mOmega\sum\nolimits_{k=1}^K\mP_k\T(\vz_k^{t}-\eta\vw_k^{t})\right),\\
\label{eq:mix-v-update}
\mW_k^{t+1}&={\rm proj}_{\gamma_k}(\mP_k\mOmega\T\valpha^{t+1}+\vz_k^t/\eta),\\
 \intertext{where the projection operator ${\rm proj}_{\lambda}$
 is the projection onto a radius $\lambda$-spectral-norm ball, as follows:}
{\rm proj}_{\lambda}(\vw)&:=\mU\min(\mS,\lambda)\mV\T,\nonumber
\end{align}
where $\mW=\mU\mS\mV\T$ is the SVD of the matricization of the input vector $\vw$.
Moreover, combining the two steps \eqref{eq:mix-v-update} and
\eqref{eq:mix-x-update-pre}, we have (see \cite{TomSuzSug09})
\begin{align}
\label{eq:mix-x-update}
\vz_k^{t+1}&=\prox{\gamma_k\eta}^{\rm tr}\left(\vz_k^{t}+\eta\mP_k\mOmega\T\valpha^{t+1}\right).
\end{align}
Note that we recover the spectral soft-threshold operation $\prox{\gamma_k\eta}^{\rm
tr}$ by combining the two steps.
Therefore, we can simply iterate steps \eqref{eq:mix-alpha-update} and
\eqref{eq:mix-x-update} (note that the term $\eta\vw_k^{t}$ in
\eqref{eq:mix-alpha-update} can be computed from
\eqref{eq:mix-x-update-pre} as $\eta\vw_k^{t}=\vz_k^{t-1}+\eta\mP_k\mOmega\T\valpha^{t}-\vz_k^{t}$.)

In order to see that the multiplier vector $\vz_k^{t}$ obtained in the
above steps converges to the primal solution of the mixture
formulation~\eqref{eq:mixture}, we take the derivative of the ordinary
Lagrangian function $L_{0}$ with respect to $\valpha$ and $\mW_k$
($k=1,\ldots,K$) and obtain the following optimality conditions:
\begin{align*}
 \valpha&=\frac{1}{\lambda}\left(\vy-\mOmega\sum\nolimits_{k=1}^{K}\mP_k\T\vz_k\right),\\
  \mP_k\mOmega\T\valpha&\in\partial \gamma_k\|\mZ_k\|_{\ast}\qquad(k=1,\ldots,K),
\end{align*}
where we used the relationship $\vw_k=\mP_k\mOmega\T\valpha$, and the
fact that $\partial\delta_{\gamma_k}(\mW_k)\ni\vz_k$ implies
$\vw_k\in\partial\gamma_k\|\mZ_k\|_{\ast}$ because the two functions
$\delta_{\gamma_k}$ and $\gamma_k\|\cdot\|_{\ast}$ are conjugate to each
other; see \cite[Cor. 23.5.1]{Roc70}. By combining the above
two equations, we obtain the optimality condition for the mixture
formulation~\eqref{eq:mixture} as follows:
\begin{align*}
-\frac{1}{\lambda}\mP_k\mOmega\T\left(\vy-\mOmega\sum\nolimits_{k=1}^{K}\mP_k\T\vz_k\right)+\partial
 \gamma_k\|\mZ_k\|_{\ast}\ni 0\qquad(k=1,\ldots,K).
\end{align*}

As in the
previous two subsections, we require that the
algorithm~\eqref{eq:mix-x-update-pre}-\eqref{eq:mix-x-update} is
invariant to scalar multiplication of the input $\vy$ and the
regularization constant $\lambda$ by the same constant $c$.
Since $\vz_k^{t}$ appears in the final solution, $\vz_k^{t}$ must scale
linearly with respect to $c$. Thus from \eqref{eq:mix-x-update-pre}, if $\valpha_k^t$ and
$\vw_k^t$ are constants with respect to $c$, the step-size $\eta$ must
scale linearly. In fact, from \eqref{eq:mix-alpha-update} and
\eqref{eq:mix-v-update}, we can see that these two dual variables remain
constant when $\vy$, $\vz_k^{t}$, and $\eta$ are multiplied by
$c$. Therefore, we choose $\eta={\rm std}(\vy)/\eta_0$.


\section{Numerical experiments}
\label{sec:exp}
In this section, we first present results on two synthetic
datasets. Finally we apply the proposed methods to the  Amino acid
fluorescence data published by Bro and Andersson~\cite{Bro97}.

\subsection{Synthetic experiments}
We randomly generated a rank-(7,8,9) tensor of dimensions (50,50,20) by
drawing the core from the standard normal distribution and multiplying
its each mode by an orthonormal factor randomly drawn from the Haar measure.
We randomly selected some elements of the true tensor for training and
kept the remaining elements for testing.
We used the algorithms described in the previous section with the
tolerance $\epsilon=10^{-3}$. We choose $\gamma_k=1$
 for  simplicity in the later two approaches. The step-size $\eta$ is
 chosen as $\eta=\eta_0/{\rm std}(\vy)$ for the first two approaches and
 $\eta={\rm std}(\vy)/\eta_0$ for the third approach with $\eta_0=0.1$.
 For the first two
 approaches,  $\lambda\rightarrow 0$ (zero observation error) was used; see
 \eqref{eq:mat-x-update-lambda=0}.  For the last approach, we used $\lambda=0$.
 The Tucker decomposition
algorithm \texttt{tucker} from the $N$-way toolbox~\cite{AndBro00} is also
included as a baseline, for which we used the correct rank (``exact'')
and the 20\% higher rank (``large''). Note that all proposed approaches
can find the rank automatically. The generalization error is defined as
follows:
\begin{align*}
 {\rm error}&=\frac{\|\vy_{\rm pred}-\vy_{\rm test}\|}{\|\vy_{\rm test}\|},
\end{align*}
where $\vy_{\rm test}$ is the vectorization of the unobserved entries
and $\vy_{\rm pred}$ is the prediction computed by the algorithms.
 For the ``As a Matrix''
strategy, error for each mode is reported.  
All algorithms were implemented in MATLAB and ran on a computer with two
 3.5GHz Xeon processors and 32GB of RAM.  The experiment was repeated 20 times and averaged.

\begin{figure}[tb]
 \begin{center}
  \includegraphics[width=.9\textwidth]{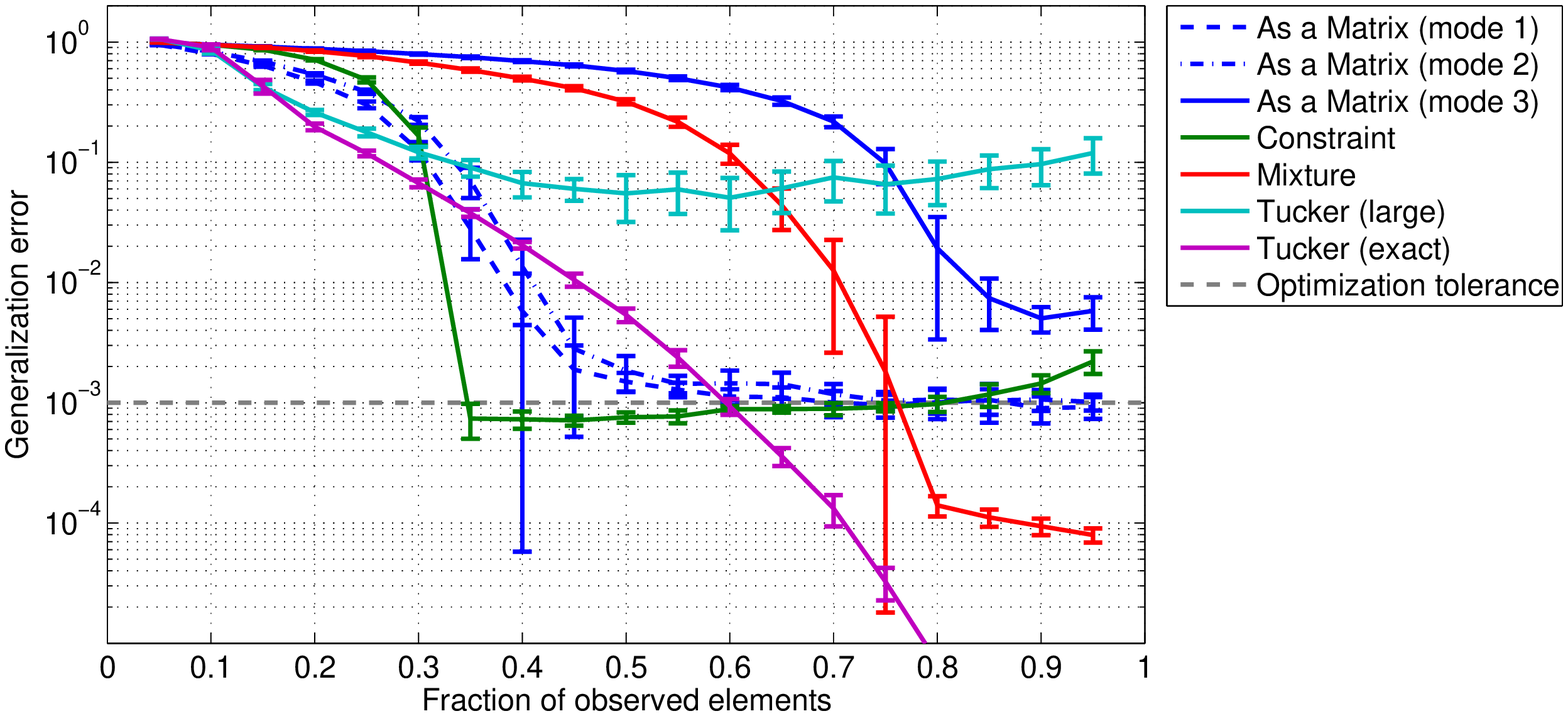}
  \caption{Comparison of three strategies, tensor as a matrix (``As a
  Matrix''), constrained optimization (``Constraint''), and mixture of
  low-rank tensors (``Mixture'') on a synthetic rank-$(7,8,9)$ tensor
  (the dimensions are $50\times 50\times 20$). Also the Tucker decomposition
  with 20\% higher rank (``large'') and with the correct rank (``exact'') implemented in the $N$-way toolbox~\cite{AndBro00} are included as 
  baselines. The generalization error is plotted against the fraction of
  observed elements of the underlying low-rank tensor. Also the
  tolerance of optimization ($10^{-3}$) is shown.}  
  \label{fig:gen}
 \end{center}
\end{figure}
\begin{figure}[tb]
 \begin{center}
  \includegraphics[width=.8\textwidth]{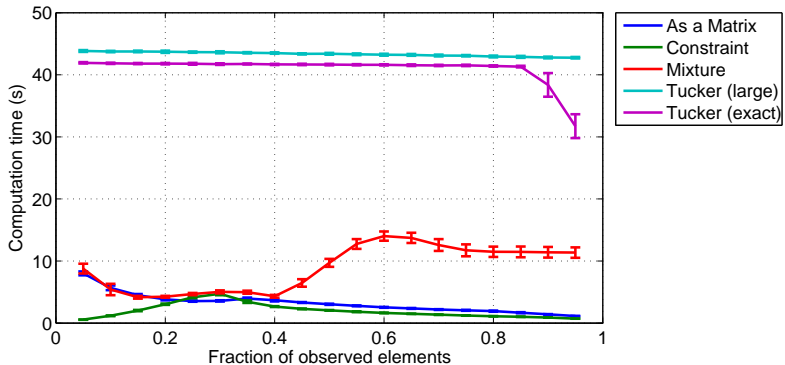}
  \caption{Comparison of computation times.}
  \label{fig:time}
\end{center}
\end{figure}
\begin{figure}[tb]
 \begin{center}
  \includegraphics[width=.45\textwidth]{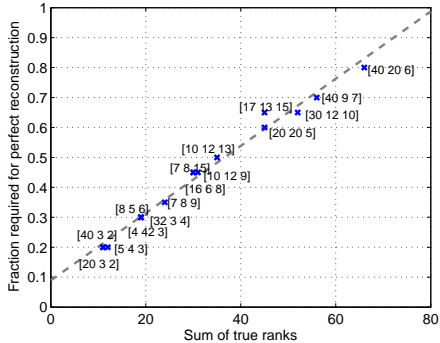}
  \caption{Fraction of observations at the threshold plotted against the
  sum of true ranks. Numbers in the brackets denote the $k$-rank of the
  underlying tensor. The dimension of the tensor is (50,50,20).}
  \label{fig:frac}
 \end{center}
\end{figure}
Figure~\ref{fig:gen} shows the result of tensor completion using three
strategies we proposed above, as well as the Tucker decomposition.
At 35\% observation, the proposed ``Constraint'' obtains nearly perfect
generalization. Interestingly there is a sharp transition from a poor fit
(generalization error$>1$) to an almost perfect fit (generalization
error$\simeq 10^{-3}$). The ``As a Matrix'' approach also show similar
transition for mode 1 and mode 2 (around 40\%), and mode~3
(around 80\%), but even the first transition is slower than the
``Constraint'' approach. The ``Mixture'' approach shows a
 transition around 70\% slightly faster than the mode~3 in the ``As A
 Matrix'' approach.
Tucker shows early
decrease in the generalization error, but when the rank is misspecified
(``large''), the error remains almost constant; even when the correct
rank is known (``exact''), the convergence is slower than the proposed
``Constraint'' approach. 

The proposed convex approaches are not only accurate but also
fast. \Figref{fig:time} shows the computation time of the proposed
approaches and EM-based Tucker decomposition against the fraction of
observed entries. For the ``As a Matrix'' approach the total time for 
all modes is plotted. We can see that the ``As a Matrix'' and
``Constraint'' approaches are roughly 4--10 times faster than the
conventional EM-based tucker decomposition. 

We have further investigated the condition for the threshold behaviour
using the proposed ``Constraint'' approach.  Here we generated different
problems of different core dimensions $(r_1,r_2,r_3)$. The sum of
mode-$k$ ranks is defined as
$\min(r_1,r_2r_3)+\min(r_2,r_3r_1)+\min(r_3,r_1r_2)$. For each problem,
we apply the ``Constraint'' approach for increasingly large fraction of
observations and determine when the generalization error falls below
$0.01$. \Figref{fig:frac} shows the fraction of observations required to
obtain generalization error below $0.01$ (in other words, the fraction
at the threshold) against the sum of mode-$k$
ranks defined above.
We can see that  the fraction at the threshold is roughly proportional
to the sum of the mode-$k$ ranks of the underlying tensor. We do not have any
theoretical argument to support this observation. Acar et
al~\cite{AcaDunKolMor10} also empirically discussed condition for
successful recovery for the CP decomposition.

\begin{figure}[tb]
 \begin{center}
  \includegraphics[width=.7\textwidth]{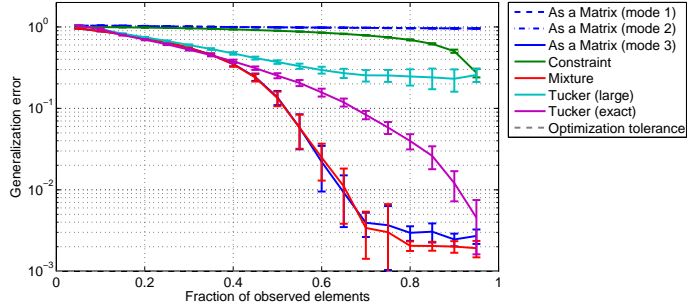}
  \caption{Synthetic experiment on a rank-$(50,50,5)$ tensor of
  dimensions $50\times 50\times 20$. See also \Figref{fig:gen}.}
  \label{fig:gen2}
 \end{center}
\end{figure}

Figure~\ref{fig:gen2} show another synthetic experiment. We randomly
generated a rank-$(50,50,5)$ tensor of the same dimensions as above. We chose the
same parameter values $\gamma_k=1$, $\lambda\rightarrow 0$,
$\epsilon=10^{-3}$, and $\eta_0=0.1$.
Here we can see that interestingly the ``Constraint'' approach perform poorly,
whereas the ``mode 3'' and ``Mixture'' perform clearly better than other
algorithms. It is natural that the ``mode 3'' approach works well
because the true tensor is only low-rank in the third mode. In contrast,
the ``Mixture'' approach can automatically detect the rank-deficient mode, because the regularization term in the
formulation~\eqref{eq:mixture} is a linear sum of three ($K=3$) penalty
terms. The linear sum structure enforces sparsity across
$\mZ_k$. Therefore, in this case $\mZ_1$ and $\mZ_2$ were switched off,
and ``Mixture'' approach yielded almost identical results to the ``mode
3'' approach.

\subsection{Amino acid fluorescence data}
\label{sec:exp-amino}
The amino acid fluorescence data is a semi-realistic data contributed by
Bro and Andersson~\cite{Bro97}, in which they measured the fluorescence
of five laboratory-made solutions that each contain different amounts of
tyrosine, tryptophan and phenylalanine. Since the ``factors'' are known
to be the three amino acids, this is a perfect data for testing whether
the proposed method can automatically find those factors. 

For the experiments in this subsection, we chose the same parameter
setting $\gamma_k=1$, $\lambda\rightarrow 0$, $\epsilon=10^{-3}$, and
$\eta_0=0.1$. Setting $\lambda\rightarrow 0$ corresponds to assuming no
observational noise. This can be justified by the fact that the original
data is already approximately low-rank (rank-$(3,3,3)$) in the sense of Tucker
decomposition. The dimensionality of the original tensor is $201\times
61\times 5$, which correspond to emission wavelength (250--450 nm),
excitation wavelength (240--300 nm), and samples, respectively.

\begin{figure}[tb]
 \begin{center}
  \includegraphics[width=.8\textwidth]{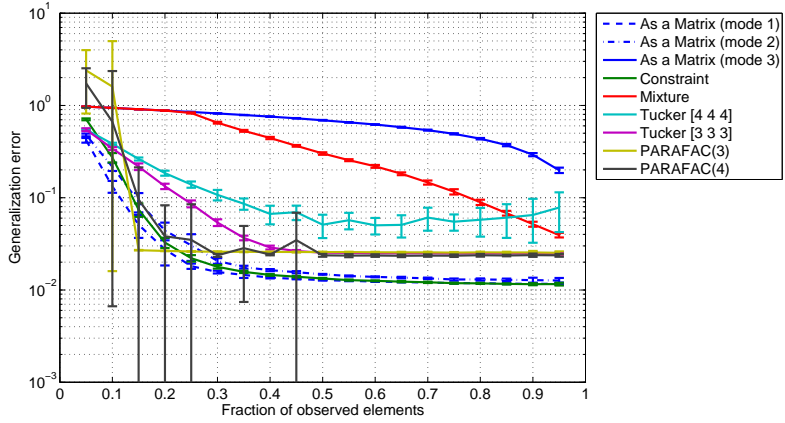}
  \caption{Generalization performance of proposed methods on the amino
  acid fluorescence data is compared to conventional EM-based Tucker
  decomposition and PARAFAC. See also Figures \ref{fig:gen} and \ref{fig:gen2}.}
  \label{fig:amino}
 \end{center}
\end{figure}

\Figref{fig:amino} show the generalization error obtained by the
proposed approaches as well as EM-based Tucker and PARAFAC decompositions.
Here PARAFAC is included because the dataset is originally designed for
PARAFAC. We can see that the proposed ``Constraint'' approach show
fast decrease in generalization error, which is comparable to the
PARAFAC model knowing the correct dimension. Tucker decomposition of
rank-$(3,3,3)$ performs as good as PARAFAC models when more than half
the entries are observed. However, a slightly larger rank-$(4,4,4)$
Tucker decomposition could not decrease the error below $0.05$.

\begin{figure}[tb]
 \begin{center}
  \includegraphics[width=.8\textwidth]{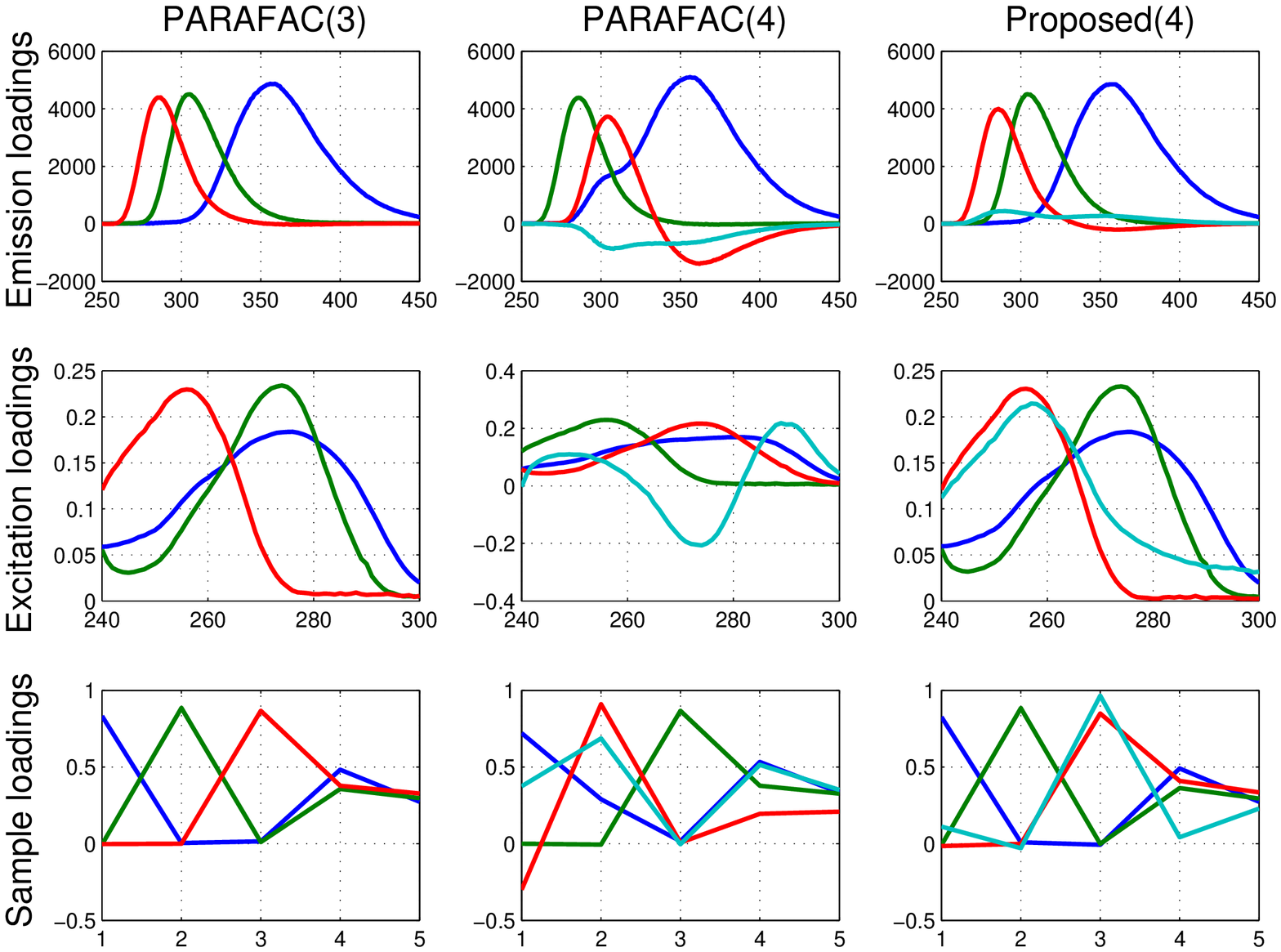}
  \caption{Factors obtained by three-component PARAFAC (left),
  four-component PARAFAC (center), and the heuristic proposed in
  \Secref{sec:interpretation} (right) at the fraction of observation
  $0.5$. Even when a redundant four-component PARAFAC is used in the
  post-processing, the proposed heuristic estimates the factors more
  reliably than directly applying the PARAFAC model.}
  \label{fig:factors}
 \end{center}
\end{figure}

\Figref{fig:factors} show the factors obtained by fitting directly three-component PARAFAC model, four-component PARAFAC model, and applying a
four component PARAFAC model to the core obtained by the proposed
``Constraint'' approach. The fraction of observed entries was $0.5$. The
two conventional approaches used EM iteration for the estimation of
missing values.  For the proposed model, the dimensionality of
the core was $4\times 4\times 5$; this was obtained by keeping the
singular-values of the auxiliary variable $\mZ_k$ that are larger than
1\% of its largest singular-value for each $k=1,\ldots,K$. Then we
applied a four-component (fully-observed) PARAFAC model to this core and obtained the
factors as in \Eqref{eq:parafac-factors}. Interestingly, although the
four component-PARAFAC model is redundant for this problem~\cite{Bro97},
the proposed approach seem to be more robust than applying
four-component PARAFAC directly to the data. We can see that the shape
of the major three components (blue, green, red) obtained by the
proposed approach (the right column) are more similar to the
three-component PARAFAC model (the left column) than the four-component
PARAFAC model (the center column).

\section{Summary}
\label{sec:summary}
In this paper we have proposed three strategies to extend the framework
of trace norm regularization to the estimation of partially observed
low-rank tensors. The proposed approaches are formulated in convex
optimization problems and the rank of the tensor decomposition is
automatically determined through the optimization.

In the simulated experiment, tensor completion using the ``Constraint''
approach showed nearly perfect reconstruction from only 35\%
observations.  The proposed approach shows a sharp threshold behaviour
and we have empirically found that the fraction of samples at the
threshold is roughly proportional to the sum of mode-$k$ ranks of the underlying tensor.

We have also shown the weakness of the ``Constraint'' approach. When the
unknown tensor is only low-rank in certain mode, the assumption that the
tensor is low-rank in every mode, which underlies the ``Constraint''
approach, is too strong. We have demonstrated that the ``Mixture''
approach is more effective in this case. The ``Mixture'' approach
can automatically detect the rank-deficient mode and lead to better
performance.

In the amino acid fluorescence dataset, we have shown that the proposed
``Constraint'' approach outperforms conventional EM-based Tucker
decomposition and is comparable to PARAFAC model with the correct number
of components. Moreover, we have demonstrated a simple heuristic to
obtain a PARAFAC-style decomposition from the decomposition obtained by
the proposed method. Moreover, we have shown that the proposed heuristic
can reliably recover the true factors even when the number of PARAFAC
factors is misspecified. 

The proposed approaches can be extended
in many ways. For example, it would be important to handle non-Gaussian
noise model~\cite{ChiKol10,HayTakShiKamKatKunYamIke10}; for example a tensor version of robust
PCA~\cite{CanLiMaWri09} would be highly desirable. For classification
over tensors, extension of the approach in \cite{TomAih07} would be
meaningful in applications including brain-computer 
interface; see  also~\cite{SigDeLSuy10b} for another recent
approach. It is also important to extend the proposed approach to handle
large scales tensors that cannot be kept in the RAM. Combination of the
first-order optimization proposed by Acar et al.~\cite{AcaDunKolMor10} with our
approach is a promising direction.
Moreover, in order to understand the threshold behaviour,
further theoretical analysis is necessary.

\appendix
\section{Computation of the dual objectives}
\label{sec:gap}
In this Appendix, we show how we compute the dual objective values for
the computation of the relative duality gap \eqref{eq:criterion}.
\subsection{Computation of dual objective for the ``As a Matrix'' approach}
The dual problem of the constrained minimization
problem~\eqref{eq:mat-const} can be written as follows:
\begin{align}
\label{eq:mat-dual-obj}
\maximize_{\valpha\in\RR^N}\quad&
 -\frac{\lambda}{2}\left\|\mOmega\mP_k\T\valpha\right\|^2+\vy\T\mOmega\mP_k\T\valpha\\
\label{eq:mat-dual-const}
\subjectto\quad&\bar{\mOmega}\mP_k\T\valpha=\bm{0},\quad\|\mA\|\leq 1.
\end{align}
Here $\bar{\mOmega}:\RR^{N}\rightarrow\RR^{N-M}$ is the linear operator
that reshapes the elements of a given $N$ dimensional vector that correspond to the unobserved
entries into an $N-M$ dimensional vector. In addition,
$\mA\in\RR^{n_k\times\nop{k}}$ is the matricization of $\valpha$ and
$\|\cdot\|$ is the spectral norm (maximum singular value).

Note that the multiplier vector $\valpha^{t}$ obtained through ADMM does not
satisfy the above two constraints~\eqref{eq:mat-dual-const}. Therefore, similar to the approach
used in \cite{WriNowFig09,TomSuzSug09}, we apply the following
transformations. First, we compute the projection $\hat{\valpha}^{t}$ by
projecting $\valpha^{t}$ to the equality
constraint. This can be done easily by setting the elements of $\valpha^{t}$
that correspond to unobserved entries to zero. Second, we compute the
maximum singular value $\sigma_1$ of the matricization of
$\hat{\valpha}^{t}$ and shrink $\hat{\valpha}^{t}$ as follows:
\begin{align*}
 \tilde{\valpha}^{t}=\min(1,1/\sigma_1)\hat{\valpha}^{t}.
\end{align*}
Clearly this operation does not violate with the equality
constraint. Finally we substitute $\tilde{\valpha}^{t}$ into the
dual objective \eqref{eq:mat-dual-obj} to compute the relative duality
gap as in \Eqref{eq:criterion}.

\subsection{Computation of dual objective for the ``Constraint'' approach}
The dual problem of the constrained minimization
problem~\eqref{eq:const-obj} can be written as follows:
\begin{align}
\label{eq:const-dual-obj}
\maximize_{\{\valpha_k\}_{k=1}^K}\quad
 &-\frac{\lambda}{2}\left\|\mOmega\ssum_{k=1}^{K}\mP_k\T\valpha_k\right\|^2+\vy\T\mOmega\ssum_{k=1}^{K}\mP_k\T\valpha_k,\\
\subjectto\quad&\bar{\mOmega}\ssum_{k=1}^{K}\mP_k\T\valpha_k=\bm{0},\quad
 \|\mA_k\|\leq \gamma_k\,(k=1,\ldots,K).\nonumber
\end{align}
Here the anti-observation operator $\bar{\mOmega}$ is defined as in the
last subsection, and $\mA_k\in\RR^{n_k\times\nop{k}}$ is the
matricization of $\valpha_k$ ($k=1,\ldots,K$).

In order to obtain a dual feasible point from the current multiplier
vectors $\valpha_k^{t}$ ($k=1,\ldots,K$), we apply similar
transformations as in the last subsection. First, we compute the
projection to the equality constraint. This can be done by computing the
sum over $\valpha_1^{t},\ldots,\valpha_K^{t}$ for each unobserved entry. Then
the sum divided by $K$ is subtracted from each corresponding entry for $k=1,\ldots,K$. Let us denote by $\hat{\valpha}_k^{t}$ the
multiplier vectors after the projection. Next, we compute the largest
singular-values $\sigma_{k,1}=\sigma_1(\hat{\mA}_k^{t})$ where $\hat{\mA}_k^{t}$
is the matricization of the projected multiplier vector
$\hat{\valpha}_k^{t}$ for $k=1,\ldots,K$. Now in order to enforce the
inequality constraints, we define
 the {\em shrinkage factor} $c$ as follows:
\begin{align}
\label{eq:shrinkage}
 c=\min(1,\gamma_1/\sigma_{1,1},\gamma_2/\sigma_{2,1},\ldots,\gamma_K/\sigma_{K,1}).
\end{align}
Using the above shrinkage factor, we obtain a dual feasible point $\tilde{\valpha}_k^{t}$ as follows:
\begin{align*}
\tilde{\valpha}_k^{t}&=c\hat{\valpha}_k^{t}\qquad(k=1,\ldots,K).
\end{align*}
Finally, we substitute $\tilde{\valpha}_k^{t}$ into the dual
objective~\eqref{eq:const-dual-obj} to compute the relative duality
gap as in \Eqref{eq:criterion}.

\subsection{Computation of dual objective for the ``Mixture'' approach}
The dual problem of the mixture formulation is already given in
\Eqref{eq:mix-dual}. Making the implicit inequality constraints
explicit, we can rewrite this as follows:
\begin{align*}
 \maximize_{\valpha\in\RR^M}\quad
 &-\frac{\lambda}{2}\|\valpha\|^2+\valpha\T\vy,\\
\subjectto\quad &\|\mP_k\mOmega\T\valpha\|\leq \gamma_k\quad(k=1,\ldots, K).
\end{align*}
Note that the norm in the second line should be interpreted as the
spectral norm of the matricization of $\mP_k\mOmega\T\valpha$.
Although the ADMM presented in
\Secref{sec:admm-mixture} was designed to solve this dual formulation, we
did not discuss how to evaluate the dual objective. Again the dual
vector $\valpha^t$ obtained through the ADMM does not satisfy the
inequality constraints.

In order to obtain a dual feasible point, we compute the largest
singular-values $\sigma_{k,1}=\sigma_1(\mP_k\mOmega\T\valpha)$ for
$k=1,\ldots,K$. From the singular-values $\sigma_{k,1}$, we can 
compute the shrinkage factor $c$ as in
\Eqref{eq:shrinkage} in the previous subsection. Finally, a dual feasible
point can be obtained as $\tilde{\valpha}=c\valpha$, which we use for the
computation of the relative duality gap~\eqref{eq:criterion}.

\bibliographystyle{siam}
\bibliography{tensor}

\end{document}